# Image Retrieval Methods in the Dissimilarity Space

Madhu Kiran[1], Kartikey Vishnu[2], Rafael M. O. Cruz[1] Eric Granger[1]

[1]LIVIA, Dept. of Systems Engineering
École de technologie supérieure Montreal, Canada
[2]Delhi Technological University
New Delhi, India

## Abstract

Image retrieval methods rely on metric learning to train backbone feature extraction models that can extract discriminant queries and reference (gallery) feature representations for similarity matching. Although state-of-the-art accuracy has improved considerably with the advent of deep learning (DL) models trained on large datasets, image retrieval remains challenging in many real-world video analytics and surveillance applications, e.g., person re-identification. Using the Euclidean space for matching limits the performance in real-world applications due to the curse of dimensionality, overfitting, and sensitivity to noisy data.

We argue that the feature dissimilarity space is more suitable for similarity matching, and propose a dichotomy transformation to project query and reference embeddings into a single embedding in the dissimilarity space.

We also advocate for end-to-end training of a backbone and binary classification models for pair-wise matching. As opposed to comparing the distance between queries and reference embeddings, we show the benefits of classifying the single dissimilarity space embedding (as similar or dissimilar), especially when trained end-to-end. We propose a method to train the max-margin classifier together with the backbone feature extractor by applying constraints to the L2 norm of the classifier weights along with the hinge loss.

Our extensive experiments on challenging image retrieval datasets and using diverse feature extraction backbones highlight the benefits of similarity matching in the dissimilarity space. In particular, when jointly training the feature extraction backbone and regularised classifier for matching, the dissimilarity space provides a higher level of accuracy.

## Introduction

Image retrieval methods compare image data in high-dimensional space such that the distance between similar query and reference images is smaller than between dissimilar ones. Metric learning is employed to train a backbone model with pairwise losses that encourage small distances for pairs of samples from the same class, and large distances for pairs of samples from different classes. Applications of image matching include person re-identification (Hermans, Beyer, and Leibe 2017), signature verification (Hafemann, Sabourin, and Oliveira 2017b), face matching (Wen et al. 2016), and object tracking (Chen et al. 2018). Before the deep learning era, classic retrieval methods commonly addressed the matching problem with Euclidean or cosine distances or by finding better Mahalanobis distances obtained due to linear transformations (Shen, Kim, and Wang 2010).

Although able to perform linear transformations with few parameters, they are insufficient to obtain higher-order correlations in the data space. Recent breakthroughs with deep learning (DL) models show they can successfully handle complex tasks involving large amounts of data by encoding patterns or relationships in the data. One specific area where they have been taken full advantage of is producing an "embedding," where the goal is to transform data into a new format that preserves certain relationships. For example, items that are similar in some way remain close to each other in the embedding space.

Two standard loss functions for metric learning are contrastive loss (Hadsell, Chopra, and LeCun 2006a) and triplet loss (Schroff, Kalenichenko, and Philbin 2015a; Sohn 2016a), widely used to train Siamese networks. In addition, discriminative embeddings have been obtained by treating the problem as a classification problem where a classifier learns to predict the labels of the examples. Cross-entropy loss is used for learning the embedding in a classification paradigm (Boudiaf et al. 2020a). However, both of these losses have their problems. Classification loss requires a growing number of learnable parameters as the number of identities increases, most of which will be discarded after training. Contrastive approaches (Le-Khac, Healy, and Smeaton 2020) learn by comparing different training examples, and this category includes losses such as contrastive loss and triplet loss. They must capture relationships among samples in a mini-batch effective method with pairwise losses. The complexity of these methods arises from various challenges, including the quality of samples, the method of mining samples, and the difficulty level in comparing these samples.

Fig 1 shows the deep features obtained from the CUB-200 (Wah et al. 2011) dataset after PCA projection. Fig 1 (a) shows multi-class deep CNN embedding pre-trained on the CUB dataset. It can be observed that some of the features belonging to similar classes are clustered while others overlap with neighboring classes. Fig 1 (b) shows the same embeddings after dichotomy transformation(element-wise difference between in-class and between-class embeddings) clus-

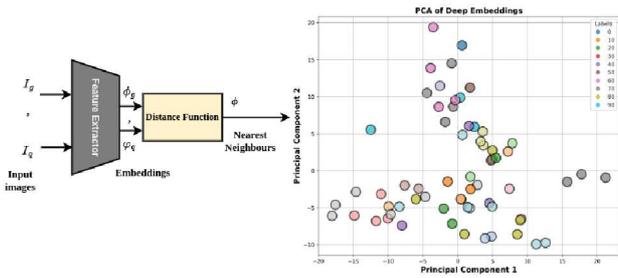

(a) Image retrieval with deep embeddings by matching the embeddings of query and gallery with a distance function

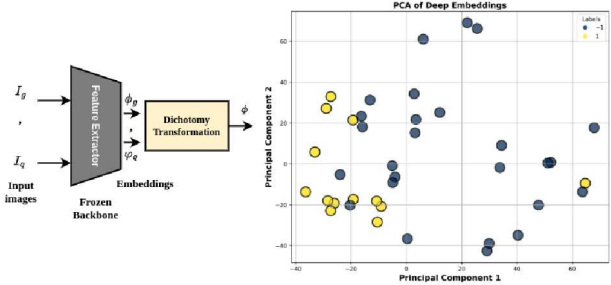

(b) Image retrieval with deep embeddings by matching the embeddings of query and gallery in the dissimilarity space

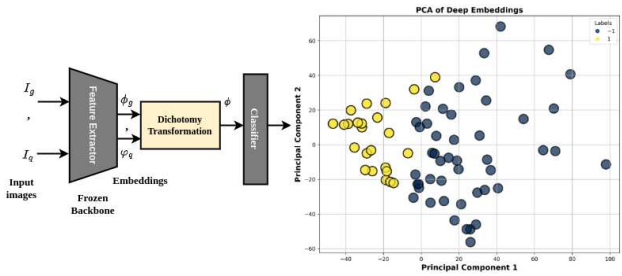

(c) Image retrieval with deep embeddings by matching the embeddings of query and gallery in the dissimilarity space trained end-to-end with a classifier for binary classification into within-class and between-class labels ter closer to the bottom left and top right, respectively. In this case, the multi-class problem is converted to a binary classification problem after dichotomy transformation. Fig 1 (c) shows embeddings after dichotomy transformation and end-to-end training with a classifier for binary classification as in-class or between-class. In this case, the classifier and feature extractor are trained end-to-end, improving the separation in the feature space. In case (b), after dichotomy transformation, the separation from the feature space is visibly improved where similar classes and dissimilar classes seem linearly separable except for a couple of outliers. In case (c), end-to-end training has further improved feature separation in the dissimilarity space.

Figure 1: An example of comparing the PCA projections of the distribution of embeddings from CUB (Wah et al. 2011) dataset obtained in deep embedding and dissimilarity space.

In addition to the aforementioned issues with contrastive learning for retrieval tasks, techniques for optimizing pairwise losses can be complex, and randomly pairing up samples often leads to slow progress or ineffective solutions (Hermans, Beyer, and Leibe 2017). More advanced techniques, such as lifted structure (Oh Song et al. 2016) and N-pair loss (Sohn 2016b), have improved sample mining to capture relationships among samples in a mini-batch effectively. However, when trained with these losses, a deep CNN maps the high-dimensional input images to low-dimensional manifolds (Hong et al. 2018; Boudiaf et al. 2020a). Euclidean distances are often used in the literature to classify a given sample in the feature space based on the proximity of a sample under the question to labeled samples in the neighborhood.

Nevertheless, from (Weller-Fahy, Borghetti, and Sodemann 2014; Xing et al. 2002), it can be observed that the Euclidean distances may suffer from the non-linearity of the manifolds. The low inter-class variations of ReID and the problem with feature space when used with Euclidean distance for classification can cause overlaps. At the same time, some of these problems can be overcome with very deep CNNs with an increasing number of parameters and better discriminative representation (Szegedy et al. 2017). Therefore, it is a trade-off between complexity and discriminatory power, and conversely, among computationally efficient deep CNNs, the above-discussed issues are augmented. Recently, many authors have explored vision transformers (He et al. 2021; Lai, Chai, and Wei 2021; Yu et al. 2022) and self-supervised pre-training for Person-ReID (Ye et al. 2022; Ma et al. 2023) and have improved the baseline by a large margin. The most common strategy to train a similarity matching network with DML includes using a deep CNN for feature extraction and combining Triplet loss (Hermans, Beyer, and Leibe 2017) and cross-entropy loss for learning similarity matching. Triplet loss for metric learning produces suboptimal separation among in-class and between-class examples (Boudiaf et al. 2020b).

Triplet loss suffers from the heuristics required for hard sample mining, and using a margin can lead to sub-optimal separation of classes (Yuan et al. 2020) if the samples are not carefully selected or can even lead to model collapsing (Ge 2018). Other loss functions, such as cross-entropy loss, can be used for learning a feature space, which could be later used for matching during test time using Euclidean distance (Zhang et al. 2020; Wang et al. 2020). Cross entropy-based feature learning strategy has once again been known to under-utilize deep feature extractors and data availability for metric learning (Hermans, Beyer, and Leibe 2017; Ming et al. 2017).

This paper introduces a new framework where similarity matching is modeled as a binary classification task, using a dichotomizer for metric learning in the dissimilarity space. In particular, the deep feature representations extracted using a pre-trained backbone are projected to the dissimilarity space, where a soft-margin classifier matches data pairs as similar or dissimilar. A classifier is trained to detect whether a given sample is obtained from a positive or negative pair. The model is trained end-to-end by optimizing the match-ing function with the feature extractor. We hypothesize that training end-to-end both the classifier and the backbone feature extractor archives better separation of class representation in the feature space. We propose a max-margin classifier to classify dissimilarity features as in-class or between-class. This is achieved by optimizing the L2 norm of the weights of the Max-Margin classifier, optimizing it for a fixed norm empirically determined as opposed to minimizing the norm without a lower limit on the norm.

This approach of training in the dissimilarity space obtains a very good performance, mainly when there are many classes and the number of examples per class is scarce (Costa et al. 2020a) in the training data. Some recent work has proposed training DL models in the dissimilarity space for biometric and surveillance applications (Souza et al. 2019, 2021), resulting in SOTA (Souza et al. 2020). In the method proposed by (Souza et al. 2019), signature verification overlaps with image retrieval is approached by converting a multi-class problem into a binary classification problem by converting embeddings of pairs of images under test to dissimilarity feature space and by using a classifier to classify them as similar or not. In addition, dissimilarity space was also used in domain adaptation to capture the relationship between between-class and within-class samples (Mekhazni et al. 2020). Our proposed model optimizes the separating hyperplane to ensure the optimal separation of classes even when training data availability is scarce, thereby improving the overall matching accuracy.

**Our main contributions are summarized as follows.**
**(1)** A framework is proposed to train a max-margin classifier in the dissimilarity space to improve matching accuracy by robust separation of classes. The deep embeddings of multi-class data are dichotomized to convert the image retrieval problem to a binary classification problem. Our proposed system performs better than conventional metric learning-based approaches by a large margin. **(2)** Our extensive experiments show the effectiveness of our proposed approach on several metric learning datasets such as Stanford Online Products (Oh Song et al. 2016), CUB (Wah et al. 2011) and Cars (Krause et al. 2013), where it can outperform conventional and state-of-the-art learning approaches. Our approach can better generalize fewer data due to the larger number of samples generated from dissimilar pairs.

## Related Work

**(1) Deep Metric Learning.** In deep metric learning, the primary objective is to learn an embedding space where data points belonging to the same class are tightly clustered. In contrast, those from distinct classes are distinctly separated. Numerous studies have introduced various loss functions to achieve this, typically falling into pair-based and proxy-based losses. Pair-based losses revolve around the interactions between pairs of data points. For instance, the contrastive loss function (Hadsell, Chopra, and LeCun 2006b), (Chopra, Hadsell, and LeCun 2005) aims to shrink the distance between positive pairs while simultaneously expanding the distance between negative pairs. Triplet loss (Weinberger and Saul 2009), (Schroff, Kalenichenko, and Philbin 2015b), (Wang et al. 2014), on the other hand, operates by considering triplets composed of an anchor, a positive sample, and a negative sample. It works towards aligning the positive pair closer than the negative pair by a predefined margin. Based on the triple loss, (Sohn 2016c) SoftMax cross-entropy was proposed to compare the group of pairs to improve pair sampling. Extensions of these loss functions have also emerged, which capture complex relationships among embedding vectors (Sohn 2016d), (Wang et al. 2019a), (Wang et al. 2019b), (Song et al. 2015). The computational cost of these pair-based works is always high due to the workload of comparing each sample with all other samples within a given batch. Additionally, these methods reveal sensitivity to the batch size, where their performance may significantly drop if it is too small. Conversely, proxy-based losses delve into the relations between data points and proxies, where a proxy serves as an additional learnable embedding vector representing a class of training data. In Proxy-NCA (Movshovitz-Attias et al. 2017), data points are associated with proxies, compelling positive pairs of a data point and its proxy to converge while concurrently pushing negative pairs apart. In Proxy-Anchor (Kim et al. 2020), each data point in a batch is connected to a proxy. This connection helps better to understand the complex relationships between the data points while depicting hardness via gradients.

**(2) Dissimilarity Space and Dichotomy Transformation.** The Dichotomy Transformation (DT), as in (Cha and Srihari 2000a), presents a methodology for reducing a $K$-class pattern recognition problem, where $K$ is large, into a binary classification problem. In metric learning, this transformation can be conceptualized as follows: given a reference image and a questioned image, the goal is to ascertain whether these two images originate from the same class or distinct classes. Post-application of the dichotomy transformation, the dissimilarity space is reduced to only two classes:

- The within-class (positive class, $u^+$): When the reference and questioned feature vectors, utilized to derive the dissimilarity vector, belong to the same class.

- The between-class (negative class, $u^-$): When the reference and questioned feature vectors, used to derive the dissimilarity vector, belong to different classes.

After the transformation into dissimilarity space, a dichotomizer (i.e., a binary classifier) is trained to perform the verification task. The trained dichotomizer is expected to effectively discern whether two samples are attributed to the same class or not (Cha and Srihari 2000a).

A key attribute of the Dichotomy Transformation is its capacity to augment the sample size within the dissimilarity space. This is achieved through the generation of pairwise comparisons between reference images. Specifically, if $K$ classes each provide $R$ reference images, the transformation can yield up to $^{KR}C_2$ distinct distance vectors, where $K\,^RC_2$ vectors pertain to the positive class and $^KC_2R^2$ to the negative class (Rivard and Granger 2011). Consequently, even with a limited number of reference images per class, the DT can produce many samples in the dissimilarity space.

This property allows to address the challenge of small sample sizes per class. Moreover, the model is better equipped to mitigate the effects of dataset imbalance by generating a balanced number of samples for both the positive (same-class) and negative (different-class) categories. Additionally, the increase in sample size enhances the model's ability to capture a broader range of intra-class variations, thereby reducing the impact of intra-class variability (Hafemann, Sabourin, and Oliveira 2017a).

**(3) Vision Transformers.** (Vaswani et al. 2017) introduced the transformer network, integrating the self-attention mechanism to address the limitations of LSTM networks (Sutskever, Vinyals, and Le 2014), (Sherstinsky 2018), notably memory degradation in handling lengthy sequences. This innovation catalyzed a wave of advancements in NLP, spawning renowned models like Bidirectional Encoder Representations from Transformers (BERT) (Devlin et al. 2018), (Liu et al. 2019), (Reimers and Gurevych 2019) and Generative Pre-trained Transformers (GPT) (Brown et al. 2020), (Radford et al. 2019). Transformers are constructed with self-attention and feedforward layers, which can be configured in various arrangements, though stacking multiple layers is the prevalent approach.

Dosovitskiy et al. (Dosovitskiy et al. 2020a) introduced the Vision Transformer (ViT) model, which resembles BERT's image classification architecture, with adaptations in the input layer. Their findings suggest that ViT when trained on sufficiently extensive datasets, performs marginally better than ResNets in tasks such as classifying Imagenet and CIFAR datasets. Touvron et al. (Touvron et al. 2020) enhanced ViT's training efficiency through knowledge distillation (Hinton, Vinyals, and Dean 2015), leading to the development of the Data-Efficient Image Transformer (DeiT). DeiT was exclusively trained on ImageNet1k, a considerably smaller dataset with 1k classes and 1.3M images (Dosovitskiy et al. 2020a), (Touvron et al. 2020). Their success hinges on knowledge distillation technique, notably surpassing conventional vanilla distillation approaches.

Another member of the ViT family, DINO (Caron et al. 2021), employs a self-supervised training paradigm. The ViT-S model is trained using the ImageNet-1k dataset without labels in this scenario. The encoder is tasked with producing consistent output for different areas of an image, achieved through augmentations like random cropping and color jitter. This training strategy aligns with image retrieval tasks, wherein the encoder is explicitly trained to yield similar output for semantically related inputs. However, the objectives of these tasks diverge: self-supervised learning gives pre-trained features for downstream tasks, whereas image retrieval relies on the resulting features directly for evaluation.

The literature on DML showcases diverse methods with distinct strengths and limitations. Pair-based losses, such as contrastive and triplet loss, have effectively created well-clustered embedding spaces. However, their computational intensity poses a significant drawback. The requirement to compare each sample with all others in a batch not only increases processing time but also makes these methods highly sensitive to batch size. Small batch sizes often lead to suboptimal performance, indicating a dependency that may not be practical in all scenarios.

Extensions to these pair-based methods have sought to address some of these issues by capturing more complex relationships among embedding vectors. While these advancements have improved performance, they often introduce additional complexity and computational overhead, hindering their widespread application.

## Proposed Method

Given a dataset $D = \{(I_i, \alpha_i)\}_{i=1}^{N}$ comprised of the original images $I_i$ and their corresponding labels $\alpha_i \in \{1, \ldots, K\}$, we aim to acquire a transformation function $h_M : D \rightarrow Y$, mapping the input data to an embedding space $Y$. We designate the embedded features set as $Y = \{y_i \in \mathbb{R}^k\}_{i=1}^{N}$.

The primary objective of DML involves enhancing the transformation function $h_M(\cdot)$, typically instantiated using a deep feature extractor as the backbone architecture. This yields features that can be evaluated with distance metrics $\delta(y_i, y_j)$ based on the similarity between sample $I_i$ and $I_j$. We hypothesize that learning the weight of the max-margin classifier with dissimilarity space is essentially learning diagonal Mahalanobis distance. Let $X$ and $Y$ be vectors or embeddings representing two samples. Then the Mahalanobis distance can be given by:

$$d_m^2(x, y) = (x - y)^\top M(x - y) \quad (1)$$

where $d_m$ represents Mahalanobis distance and $M$ is the inverse covariance matrix of the data that can be re-written as a function of $L$,

$$M = L^\top L \quad (2)$$

where $L$ is non-arbitrary and often derived from the Cholesky decomposition of the inverse covariance matrix $M$. Substituting Eqn 2 into Eqn 1:

$$d_m^2(x, y) = \|L(x - y)\|^2 \quad (3)$$

Similarly, the distance of a sample from the margin with SVM can be rewritten as:

$$d_{\text{margin}}^2 = \|W(x - y)\|^2 \quad (4)$$

Where $W$ is a diagonal square matrix with a principal diagonal representing the roots of SVM weights. Comparing Eqn 4 and Eqn 3, both can be described as a product of the difference between the features and a square matrix. Hence, the square matrix $W$ is equivalent to the diagonal Mahalanobis matrix. Thus, Learning SVM weights $W$ in the dissimilarity space is comparable to learning the diagonal Mahalanobis matrix.

Inspired by dissimilarity learning methods (Oliveira; Costa et al. 2020b), the proposed method aims to maximize the separation of dissimilar classes by end-to-end optimization of the backbone feature extractor with a linear classifier in the dissimilarity space. In addition, a simple, lightweight adapter module is used before transforming the features to dissimilarity space from Euclidean space. Our proposed metric framework is illustrated in Fig. 2.

The deep neural network backbone $f(\cdot)$ extracts representation vectors from image data samples $I$. The embedding extracted from this is represented by $\psi = f(\boldsymbol{I})$ where $\psi$ is the flattened output with the adaptive average pooling layer. Our architecture allows any deep neural network feature extractors to be used for $f(\cdot)$. For similarity matching, two input images, $\boldsymbol{I_q}$ (query) and $\boldsymbol{I_r}$ (reference), are processed by the backbone, allowing to extract feature vectors $\psi_q$ and $\psi_r$, respectively.

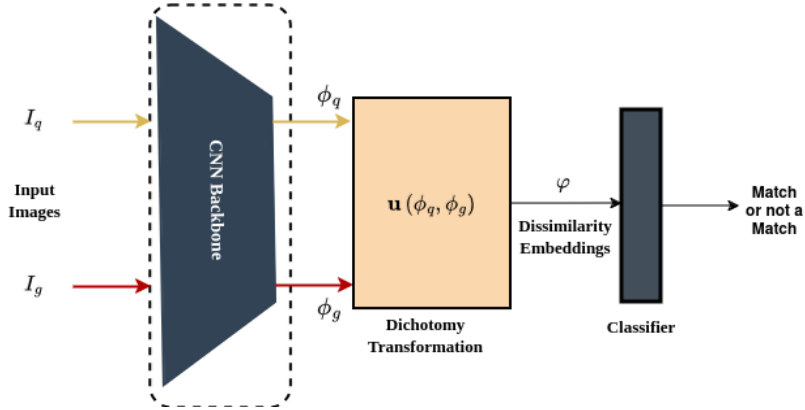

Figure 2: The proposed framework for end-to-end metric learning in the dissimilarity space. A backbone model $f(\cdot)$ extracts feature embeddings from input images that undergo dichotomy transformation to obtain a dissimilarity vector $\mathbf{u}(\phi_q, \phi_g)$. The max-margin classifier outputs the dissimilarity score and positive or negative class prediction.

In the adapter layer, a set of linear weights, one weight per element, is applied to the features $\psi$ before processing by the dichotomizer module. The backbone pre-trained network has been trained in the Euclidean space. However, the dissimilarity embedding space can be sparse, particularly the within-class examples close to the origin. Therefore, the distribution is very different from that of Euclidean space. Hence, the adapter layer helps to adapt the Euclidean to the dissimilarity feature embeddings. The adapted features $\phi$ are defined by $\phi = \sigma\left(W^{(1)}(\psi)\right)$, where $W$ are the weights, and $\sigma$ is an activation unit (in our case ReLU).

The dichotomy transformation block facilitates the conversion of a multi-class problem into a binary classification task. This process involves organizing the extracted features into pairs representing the absolute differences between the feature vectors, which is then used for binary classification. The dichotomy transformation approach, as introduced by Cha et al. (Cha and Srihari 2000b), achieves this conversion through a two-step procedure. Initially, a combination of input features is created to form pairs. These pairs are categorized as within-class pairs, originating from the same class, or between-class pairs if they differ. Subsequently, the pairs are further transformed into a single feature vector within the dissimilarity space. Let $\phi_q$ and $\phi_q$ be query and gallery feature vectors extracted from image samples $\boldsymbol{I}_q$ and $\boldsymbol{I}_g$, respectively. Then, the dissimilarity vector resulting from the dichotomy transformation can be represented by,

$$\mathbf{u}(\phi_q, \phi_g) = \begin{bmatrix} |\varphi_{q1} - \varphi_{g1}| \\ |\varphi_{q2} - \varphi_{g2}| \\ \vdots \\ |\varphi_{qn} - \varphi_{gn}| \end{bmatrix}, \quad (5)$$

Eqn. 1 calculates absolute value of the difference between $i_{th}$ and $j_{th}$ dimension of the embedding, i.e., $\varphi_{qi}$ and $\varphi_{gi}$ of the feature vectors $\phi_q$ and $\phi_g$ respectively. Dissimilarity space is expected to map a given sample obtained after dichotomy transformation close to the origin of the space if the sample is a within-class sample and farther away from the origin if the sample is a between-class sample. This enables a linear classifier like a max-margin classifier to find an optimal separating hyperplane.

A max-margin classifier $W_c(\cdot)$ classifies a given dissimilarity vector $\mathbf{u}(\phi_q, \phi_g)$ into positive or negative classes. It outputs a binary class $\mathbf{c}$ using the dissimilarity feature vector from the previous step, $n(\mathbf{u})$.

**End to End Training:** Given a mini-batch of training data, within-class samples and between-class sample pairs undergo dichotomy transformation after extracting the embeddings with the backbone. The embeddings used for the dichotomy transformation are output embeddings of the adaptive average pooling layer, meaning they are 3D tensors smaller than the feature map output from the deep neural network and the adapter layer. The backbone deep neural network(DNN) feature extractor may also be trained simultaneously with conventional triplet loss or other contrastive loss functions, as discriminative features are essential for the dissimilarity space. The backbone feature extractor training is performed on the output embeddings of the adaptive average pooling layer, CNN's global average pooling layer, and adapter layer to follow the conventional training protocol in metric learning framework (Boudiaf et al. 2020a).

The positive samples represented by $y$ are labeled 1, and negative examples are labeled $-1$. The max-margin classifier (?) is essentially a fully connected layer with weights $W_c$ trained with hinge loss. As discussed, the losses are jointly optimized along with $L_2$ norm on the weights $W_c$ of the linear classifier to train a max-margin classifier,

$$\begin{aligned} \mathcal{L}_{\text{hinge}} &= \tfrac{1}{2}\|\boldsymbol{W}_c\|^2 \\ &+ C \sum_{n=0}^{N} \max\left\{0, 1 - y_n\left(\boldsymbol{W_c}^\top \boldsymbol{u}_n + b\right)\right\} \end{aligned} \quad (6)$$

The system must be jointly optimized in the embedding and dissimilarity space as a good feature representation is important for dissimilarity space to work in practice (Duin et al. 2010; Oliveira; Costa et al. 2020b). The feature space is optimised by $\varphi$ with triplet($\mathcal{L}_{\text{tri}}$) loss (Mekhazni et al. 2020) and cross-entropy($\mathcal{L}_{\text{CE}}$) (Mekhazni et al. 2020) for ID classification. Finally, the joint loss is optimized for the convergence of the feature extractor and Max-Margin classifier.

$$\mathcal{L}_{\text{total}} = \mathcal{L}_{\text{ce}} + \mathcal{L}_{\text{tri}} + \mathcal{L}_{\text{hinge}} \quad (7)$$

---

**Algorithm 1:** End-to-end learning for dissimilarity space.

**Require:** batch of image pairs $qg_1, ..., qg_B$, of gallery size $B$
model parameters $\theta$
**Ensure:** Binary classification of input pairs $y$, trained parameters $\theta$ ($f(\cdot), W_c, W$)

$\psi_q, \psi_g \leftarrow f(q), f(g)$ ▷ Obtain embeddings
$\phi_q, \phi_g \leftarrow W(\psi_q), W(\psi_g)$ ▷ Adapted embeddings
$\mathbf{u} \leftarrow \phi_g, \phi_q$ ▷ Convert to dissimilarity vector
$y \leftarrow W_c(\mathbf{u})$ ▷ Binary classification
Compute hinge loss $L_{hinge}$ from Eqn 7
Update model parameters: $Optimize(L_{hinge}(\theta))$

---

The end-to-end optimization of the Metric framework is summarized in Algorithm 1, where the overall model parameters are represented by $\theta$. The model parameters are updated by optimizing Eqn 7.

## Results and Discussion

### Experimental Methodology

**(1) Datasets:**
The **CUB200-2011** dataset consists of 11788 images representing 200 different classes of birds. The images were resized so that their smallest side was 256 pixels long. The brightness, saturation, and contrast were randomly changed. The **CARS196** dataset consists of 16185 images, evenly distributed among 196 classes. Images are resized to 256x256; then, a patch is cropped from a random section, again resized to 224x224. The brightness, saturation, and contrast were also randomly changed. The **SOP** dataset comprises 1,0583 online product images from 22634 categories. We observed that there are about 2 - 10 images for each category. Images are resized to 256x256. Then, a patch is cropped from random sections and resized to 224x224. Compared to CUB200-2011 and CARS196, the data distribution here is significantly different. The **In-shop clothes Retrieval** dataset is a large subset of DeepFashion. It contains large pose and scale variations and large diversities, quantities, and rich annotation. It consists of 7,982 different clothing items and 52,712 different in-shop clothes images. It also includes about 200k cross-pose or scale pairs.

**(2) Implementation Details:**
Our method was implemented with various backbones without any modifications, as mentioned in the experiments below. The Max-Margin classifier was implemented with a fully connected layer trained with hinge loss and weight decay. In addition to this, the classifier was used with a Batch norm layer and dropout layer. A batch size of 128 was used for all our experiments, and the image size was fixed to 256x256 unless and until stated in the experiments below.

**(3) Data Augmentation:**
As is typical when training deep learning models, using data augmentation helps improve the methods' overall performance. For the CUB dataset, images are resized, maintaining an aspect ratio of 256 pixels. Images are resized directly to 256 × 256 pixels for the other test datasets mentioned above. After resizing, a random image patch is selected and resized to 224 × 224 pixels.

**(4) Backbone feature extractors:**
Since our proposed method is agnostic to backbone architecture, we propose to implement our method on both CNN-based and transformer architecture. For the CNN-based architecture, we employed ResNet50 (He et al. 2016) architecture, and for the transformer-based method, we employed ViT-S 16/316 (Dosovitskiy et al. 2020b) owing to their popularity in metric learning frameworks.

### Comparison with State-of-Art Methods

Tab 1 shows the results of state-of-the-art contrastive learning-based approaches compared with end-to-end learning in dissimilarity space. We implement the DeIT backbone with our end-to-end learning for dissimilarity space to compare with corresponding cosine similarity learning approaches and other approaches in the literature. Results show that end-to-end learning approaches in the dissimilarity space improve the overall retrieval accuracy and robustness. Additional results, including qualitative results, are presented in the supplementary materials.

### Ablation Studies

To improve and verify the effectiveness of our Max-margin classifier-based image retrieval against the distance-based metric, our proposed method was tested on the SOP and CUB200 datasets. A pre-trained backbone was trained on cross-entropy loss and Triplet Losses to use the embeddings with distance-based metrics for image retrieval. For a fair comparison, these pre-trained embeddings are then trained with a max-margin classifier, keeping the feature extractor frozen and training the Max-Margin classifier alone.

**(1) Classifier-based learning can help improve overall retrieval accuracy:**
We hypothesize that learning similarity or image retrieval as a classification problem helps alleviate some of the issues of pairwise contrastive learning. In particular, it helps avoid overfitting in contrastive learning. To demonstrate this, we compare the performance of various backbone feature extractors for retrieval tasks trained with contrastive learning with those of the backbones trained with dissimilarity features-based learning. For a fair comparison, the backbones learned with contrastive learning are frozen, and a Max-Margin classifier was trained on features extracted from frozen backbones on various datasets. This is shown in Tab 3. The results indicate consistent improvement in the case of dissimilarity-based learning for retrieval.

Table 1: Results of our method compared with state-of-the-art metric learning methods with contrastive learning compared with end-to-end learning based approach for dissimilarity space.

| Method | Backbone | Venue | SOP | | | | CUB200 | | | | In-Shop | | | |
|---|---|---|---|---|---|---|---|---|---|---|---|---|---|---|
| | | | 1 | 10 | 100 | 1000 | 1 | 2 | 4 | 8 | 1 | 10 | 20 | 30 |
| Margin | ResNet50 | ICCV 2019 | 72.7 | 86.2 | 93.8 | 98 | 63.9 | 75.3 | 84.4 | 90.6 | - | - | - | - |
| FastAP | ResNet50 | CVPR 2019 | 73.8 | 88 | 94.9 | 98.3 | | - | - | - | - | - | - | - |
| MIC | ResNet50 | ICCV 2019 | 77.2 | 89.4 | 94.6 | - | 66.1 | 76.8 | 85.6 | - | 88.2 | 97 | - | 98 |
| NSoftmax | ResNet50 | BMVC 2018 | 79.5 | 91.5 | 96.7 | - | 65.3 | 76.7 | 85.4 | 91.8 | 89.4 | 97.8 | 98.7 | 99 |
| ProxyNCA++ | ResNet50 | ECCV 2020 | 81.4 | 92.4 | 96.9 | 99 | 72.2 | 82 | 89.2 | 93.5 | 90.9 | 98.2 | 98.9 | 99.1 |
| UML | ResNet50 | ECCV 2020 | 81.1 | 91.7 | 96.3 | 98.8 | 69.2 | 79.2 | 86.9 | 91.6 | 90.6 | 98 | 98.6 | 98.9 |
| Diss+SVM | ResNet50 | Ours | 83.7 | 92.8 | 97.5 | 99.1 | 67.4 | 77.9 | 85.3 | 90 | 90.9 | 98.1 | 98.6 | 98.9 |
| HIER | DeIT | CVPR 2023 | 84.2 | 93.7 | 97.3 | 99.1 | 76.6 | 85 | 91.1 | 94.3 | 91.9 | 98.1 | 98.7 | 98.9 |
| Diss+SVM | DeIT | Ours | 85.1 | 93.9 | 97.8 | 99.2 | 78.5 | 86.3 | 92 | 94.8 | 92.1 | 98.2 | 98.9 | 99.1 |

Note that the datasets have been categorized as fine-grained and diverse classification problems. CUB (Wah et al. 2011) datasets have subtle differences between classes as all of them belong to the bird category, while Stanford Online-Products exhibit more diversity in the classes as they represent different products. From the table, it can be observed that our method works in both the categories of datasets by improving the baseline by 1-2%.

Table 2: Results of our method implemented with different feature extractor backbones and compared with the contrastive learning-based method. In the Table, "ours" represents a frozen feature extractor trained with a classifier.

| Backbone | Fine Grained Variations | | | | Diverse Classes and Variations | | |
|---|---|---|---|---|---|---|---|
| | CUBS | | | | Stanford Online Products | | |
| | R-1 | R-2 | R-4 | R-8 | R-1 | R-10 | R-100 |
| GoogleNet | 53.7 | 65.7 | 76.7 | 85.7 | 70.1 | 84.9 | 93.2 |
| GoogleNet (ours) | **55.2** | 66.8 | 77.9 | 87.2 | 70.7 | 85.8 | 93.7 |
| ResNet18 | 63.9 | 75.0 | 83.1 | 89.7 | - | - | - |
| ResNet18 (ours) | 64.4 | 75.9 | 83.2 | 90.0 | - | - | - |
| ResNet50 | 69.2 | 79.2 | 86.9 | 91.6 | 75.9 | 88.4 | 94.9 |
| ResNet50 (ours) | 69.4 | 80.1 | 87.2 | 92.1 | **77.2** | 90.0 | 96.6 |
| ViT-S16/384 | 86.6 | 91.7 | 94.8 | 97.7 | 84.4 | 93.6 | 97.3 |
| ViT-S16/384 (ours) | **88.3** | 93.5 | 96.0 | 99.6 | 85.0 | 93.6 | 97.6 |
| ViT-B16/512 | 88.5 | 92.8 | 95.1 | 98.4 | 88.0 | 96.1 | 98.6 |
| ViT-B16/512 (ours) | 88.9 | 93.4 | 95.6 | 99.2 | **89.6** | 97.5 | 99.6 |

**(2) Max-Margin based classifier has advantages over Mahalanobis matrix learning:**
In the previous section, we have shown that, theoretically, the weights of the Max-Margin classifier boil down to a diagonal Mahalanobis matrix. This has fewer parameters but does not consider covariance among features. Tab 4 compares the results of End-to-End learning with dissimilarity space and full Mahalanobis matrix learning. It can be observed from the results that although Mahlanobis matrix learning improves results over the baseline methods, Dissimilarity-based learning outperforms Mahalnobis matrix learning by 0.5%. This could be because of learning fewer parameters with dissimilarity space and avoiding overfitting. Also, fewer parameters leads to lower computational complexity since SVM is learnt as a linear layer and the weights being a vector equal to the number of features, as opposed to a square matrix.

Table 4: Comparison of end-to-end training with our proposed method compared with Malahanobis distance learning

| Backbone | Method | R1 | R2 | R4 | R8 |
|---|---|---|---|---|---|
| ResNet18 | Cross Entropy | 63.9 | 75 | 83.1 | 89.7 |
| | Mahalanobis | 64.17 | 74.19 | 82.6 | 90.04 |
| | Ours (End to End) | **65.2** | **76.8** | **84.1** | **90.9** |
| ViT-S16/316 | Cross Entropy | 86.6 | 91.7 | 94.8 | 97.7 |
| | Mahalanobis | 88.1 | 93.6 | 95.9 | **98.1** |
| | Ours (End to End) | **88.8** | **94** | **96.2** | 98 |

| Dataset Percentage | Diss+SVM (Ours) | Euclidean | Delta |
|---|---|---|---|
| 100 % | 88.71 | 84.76 | +3.95 |
| 50 % | 77.53 | 72.11 | +5.42 |
| 25 % | 73.56 | 70.4 | +3.16 |

Table 5: Impact of data size on R-1 retrieval accuracy. The Delta column shows the difference between Diss+SVM (Ours) and Euclidean.

**(3) Impact of data size:**
We advocate that one of the main advantages of dissimilarity space with a binary classifier is the availability of several examples generated from fewer training samples to train the classifier. In additions, the Max-Margin classifier does not overfit the large number of samples generated mainly due to the norm on the weights, and the Max-Margin classifier allows for a few miss-classifications. To verify this claim, we experiment with the CUBS200 dataset, where we downsample the dataset to 50%, 25% the dataset size, and train the system on both Euclidean distance learning setup and Dissimilarity space with SVM. From Tab 5, it can be observed that dissimilarity space with SVM is robust to fewer data than Euclidean space. In particular, it can be observed that the results of 25% train data with dissimilarity space performance are better than 50% train data with Euclidean space.

## Conclusion

In this paper, a novel method for end-to-end learning of matching problems from image retrieval applications has been presented, which uses the dissimilarity transformation and classifier base approach to classify pairs of samples as similar or dissimilar as opposed to Euclidean distance or cosine similarity-based pairwise matching. When used with a dichotomizer, we argue that dissimilarity transformation converts a multi-class problem to a binary classification problem, alleviating the problem of distance-based learning with contrastive loss functions. In addition, dissimilarity-based learning techniques with a Max-Margin classifier alleviate the problem of overfitting that can be introduced in distance-based learning methods. To this end, we achieve consistent improvements not just while using a classifier as a replacement for distance matching but also better improvement when learning with a classifier end-to-end. The proposed method was implemented with different backbones and tested on metric learning datasets. The results show that having 50% of data can produce similar results in the dissimilarity space compared to that of trained with 100% data in the feature space. Additional ablation studies were performed with SVM, such as optimization with frozen deep features and deep features with end-to-end learning. These ablation studies show that optimizing the SVM-like classifier alone on deep features improves retrieval results, and optimizing features and the classifier together improves the overall results by 2-3%.